\documentclass[10pt,twocolumn,letterpaper]{article}

\usepackage{wacv}
\usepackage{times}
\usepackage{epsfig}
\usepackage{graphicx}
\usepackage{amsmath}
\usepackage{amssymb}
\usepackage{booktabs}
\usepackage{subcaption}
\usepackage{enumitem}
\usepackage[accsupp]{axessibility}
\usepackage{stfloats}

\def\wacvPaperID{996} %

\wacvfinalcopy %

\ifwacvfinal
\usepackage[breaklinks=true,bookmarks=false]{hyperref}
\else
\usepackage[pagebackref=true,breaklinks=true,colorlinks,bookmarks=false]{hyperref}
\fi

\begin{document}

\title{Structure-Encoding Auxiliary Tasks for Improved Visual Representation in Vision-and-Language Navigation}
\author{Chia-Wen Kuo\\
Georgia Tech\\
{\tt\small albert.cwkuo@gatech.edu}
\and
Chih-Yao Ma\\
Georgia Tech\\
{\tt\small cyma@gatech.edu}
\and
Judy Hoffman\\
Georgia Tech\\
{\tt\small judy@gatech.edu}
\and
Zsolt Kira\\
Georgia Tech\\
{\tt\small zkira@gatech.edu}
}
\maketitle
\thispagestyle{empty}

\begin{abstract}
In Vision-and-Language Navigation (VLN), researchers typically take an image encoder pre-trained on ImageNet without fine-tuning on the environments that the agent will be trained or tested on.
However, the distribution shift between the training images from ImageNet and the views in the navigation environments may render the ImageNet pre-trained image encoder suboptimal.
Therefore, in this paper, we design a set of structure-encoding auxiliary tasks (SEA) that leverage the data in the navigation environments to pre-train and improve the image encoder.
Specifically, we design and customize (1) 3D jigsaw, (2) traversability prediction, and (3) instance classification to pre-train the image encoder.
Through rigorous ablations, our SEA pre-trained features are shown to better encode structural information of the scenes, which ImageNet pre-trained features fail to properly encode but is crucial for the target navigation task.
The SEA pre-trained features can be easily plugged into existing VLN agents without any tuning.
For example, on Test-Unseen environments, the VLN agents combined with our SEA pre-trained features achieve absolute success rate improvement of 12\% for Speaker-Follower~\cite{fried2018speaker}, 5\% for Env-Dropout~\cite{tan2019learning}, and 4\% for AuxRN~\cite{zhu2020vision}.
\end{abstract}
\section{Introduction} \label{sec:introduction}

\begin{figure*}
\centering
\includegraphics[width=1.0\textwidth]{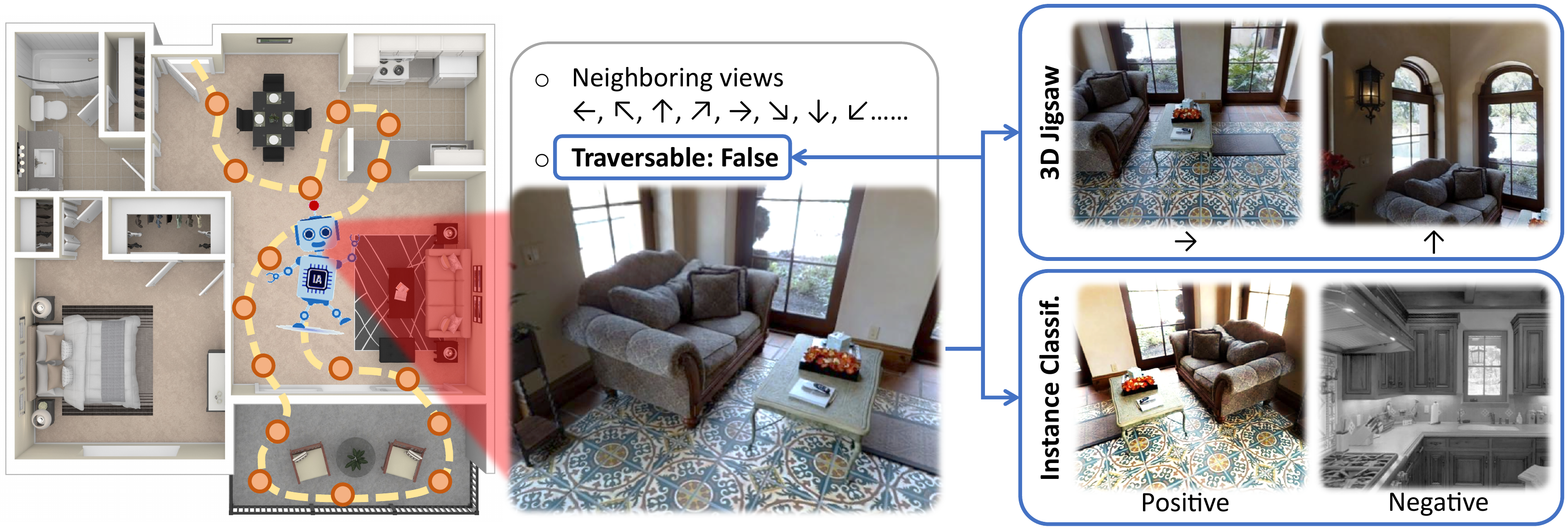}
\caption{
We propose three auxiliary tasks: (1) 3D Jigsaw, (2) Traversability Prediction, and (3) Instance Classification,
to improve the visual representation for the downstream VLN task.
These auxiliary tasks train only on data available in the navigation environments such as RGB image views, a view's neighboring views, and traversable directions within a view.
}
\label{fig:concept}
\end{figure*}

In Vision-and-Language Navigation (VLN)~\cite{anderson2018vision}, an agent navigates in a complicated environment to a target location by following human instructions.
In this task, the agent needs to interpret human instructions, encode visual input, and then infer appropriate actions according to the joint textual and visual information.
Remarkable progress has been made since the proposition of the Room-to-Room (R2R) dataset by Anderson et al.~\cite{anderson2018vision}, including generating more training data~\cite{fried2018speaker,sotp2019acl,tan2019learning}, learning a better joint visual and textual representation~\cite{wang2019reinforced,huang2019transferable,hao2020towards,majumdar2020improving}, improving the agent's internal state representation for the \textit{policy} network  (as opposed to visual encoder)
 by auxiliary tasks~\cite{ma2019selfmonitoring,zhu2020vision,wang2020environment}, and so on.

However, most of the existing works ignore the importance of the underlying visual representation by simply taking an image encoder (CNN model that encodes an image $x$ into a feature $f_x \in R^d$) pre-trained on ImageNet to encode the views in the navigation environments (\textit{e.g.,} Matterport3D~\cite{Matterport3D}).
Because of the data distribution shift between ImageNet and the navigation environments, as well as the difference between the pre-training task (image classification) and the target task (VLN), the ImageNet pre-trained image encoder may not be able to encode information crucial for the VLN task.
One na\"ive way to mitigate this negative effect is to fine-tune the image encoder on the target environments and task.
However, in the navigation environments, image labels such as semantic segmentation masks, object bounding boxes, or object and scene classes may not be available for fine-tuning the image encoder.
Furthermore, it is computationally prohibitive\footnote{To train an agent with panoramic action space, in each iteration, we take 64 trajectories, each trajectory contains 5 steps on average, each step contains 36 views, and each view is a $640\times480$ high-resolution image.
These sum up to 10k+ forward passes of high-resolution images through the image encoder in just one training iteration.
}
to fine-tune the image encoder jointly with the agent on the target VLN task.

To improve the image encoder without the need for manually annotated labels
in the target environments and without fine-tuning with the VLN agent jointly, we pre-train the image encoder on proposed \textit{structure-encoding auxiliary tasks (SEA)} with data available in the navigation environments shown in Figure~\ref{fig:concept}. 
Specifically, we collect RGB images from different views of the environments, a view's neighboring views, and traversable directions within a view.
After that, we pre-train the image encoder via the proposed auxiliary tasks on the gathered data.
We then pre-compute the features for each view of the training environments using the frozen, pre-trained image encoder, and train the navigation agent following the classical VLN methods with our pre-computed features.

Combined with our SEA pre-trained features, VLN methods achieve absolute success rate improvement of 12\% for Speaker-Follower~\cite{fried2018speaker}, 5\% for Env-Dropout~\cite{tan2019learning}, and 4\% for AuxRN~\cite{zhu2020vision} on Test-Unseen under the single-run setting (\textit{i.e.,} without pre-exploration or beam search).
To understand how the agents benefit from our SEA pre-trained features, we conduct thorough ablation studies on what information is encoded by and how the agent's navigation performance is affected by each auxiliary task.
Compared with ImageNet pre-trained features, our SEA pre-trained features better encode structural information of the scenes, which are crucial for the target navigation tasks.
The source code and collected data for pre-training the image encoder as well as the pre-trained SEA features will be released to facilitate future research in VLN.
Our key contributions are that we:
\begin{itemize}[topsep=0pt,itemsep=-1ex,partopsep=1ex,parsep=1ex,labelindent=0.0em,labelsep=0.2cm,leftmargin=*]
    \item Design and customize a set of auxiliary tasks to improve the visual representation by training on images and metadata easily attainable in the navigation environments.
    \item Achieve significant performance improvement on the unseen environments when combining our SEA pre-trained features with VLN methods including Speaker-Follower~\cite{fried2018speaker}, Env-Dropout~\cite{tan2019learning}, and AuxRN~\cite{zhu2020vision}.
    \item Conduct thorough ablation studies to understand how an agent benefits the proposed auxiliary tasks and SEA pre-trained features.
\end{itemize}

\section{Related Works}

\subsection{Auxiliary Tasks for Training an Agent}

In vision and language navigation, due to the limited amount of training data, researchers have proposed auxiliary tasks to regularize the agent model and to provide additional training signals. Despite their success in the VLN task, this line of works focuses on refining the agent's internal state representation for the policy network rather than the visual representation, and simply encode each view with a frozen, ImageNet~\cite{ILSVRC15} pre-trained image encoder.
Ma et al.~\cite{ma2019selfmonitoring} proposed a progress monitoring module to improve the grounding between visual and textual information.
Huang et al.~\cite{huang2019transferable} also aimed at improving the grounding between visual and textual information by a cross-modal alignment loss to classify aligned instruction-path pairs. To further improve the agent's state representation, they propose a coherence loss to predict future $k$ steps.
Another way of improving the cross-modal grounding is to pre-train the model on paired data of image and text \cite{hao2020towards,majumdar2020improving}.
To improve the model's generalizability, Wang et al.~\cite{wang2020environment} proposed an adversarial training strategy to remove scene-specific information from the agent's state representation.
Zhu et al.~\cite{zhu2020vision} achieved significant improvement over the previous state of the art with four auxiliary tasks including speaker model, progress monitor, orientation prediction, and trajectory-instruction matching.

In reinforcement learning (RL), it has been shown that jointly training the agent with auxiliary tasks improves state representations and greatly expedites training.
Some common auxiliary tasks in RL include: (1) future prediction~\cite{lee2020predictive,gordon2019splitnet,ye2020auxiliary,DK2017,pathak2017curiosity}, which predicts an agent's future state conditioned on its current state and the actions taken, (2) inverse dynamic~\cite{gordon2019splitnet,ye2020auxiliary,pathak2017curiosity}, which predicts the actions taken between two states, and (3) contrastive learning~\cite{anand2019unsupervised,srinivas2020curl}, which applies contrastive learning to refine the state representation.
Despite differences between RL works and VLN, AuxRN~\cite{zhu2020vision} (which we further improve upon) incorporate an auxiliary task of agent's orientation prediction in VLN, which is similar to the inverse dynamic auxiliary task in RL.
We also draw inspiration from the RL line of works to propose our auxiliary tasks.
For example, the concept of traversability prediction is similar to Chaplot et al.~\cite{chaplot2020neural}, which target building a topological map of the environment for image-goal navigation.
Different from these lines of work, our proposed auxiliary tasks focuses on the improvement of visual representation, which is later used by a VLN agent for navigation and state representation.
Our proposed auxiliary tasks are effective and improve upon AuxRN, which introduces several auxiliary tasks for training the VLN policy, by large margins.

\subsection{Self-Supervised Learning in Computer Vision}

Self-supervised learning has achieved great success in learning good visual representations without labels for data.
The learned representations generalize well to a wide variety of downstream tasks such as image classification, object detection, scene classification, and so on.
To train the model without labels, self-supervised learning methods define auxiliary tasks by visual clues that are inherent to the data.
Such visual clues include: spatial information from images~\cite{doersch2015unsupervised,noroozi2016unsupervised}, spatial-temporal information from video or motion~\cite{wang2015unsupervised,agrawal2015learning,pathak2017learning}, image colors~\cite{pathak2016context,zhang2016colorful,larsson2016learning,larsson2017colorization}, etc.
Recently, contrastive learning~\cite{oord2018representation,tian2019contrastive,chen2020simple,he2020momentum,grill2020bootstrap,caron2020unsupervised} has achieved comparable performance with its supervised counterpart.
The goal of contrastive learning is to learn a visual representation that is invariant to a set of image augmentations~\cite{tian2020makes,chen2020simple,zhao2020makes,zoph2020rethinking,xiao2020should} by identifying an image's augmented copy from a pool of other images.

Inspired by the success of self-supervised learning in computer vision, in this work we design three auxiliary tasks to improve the image encoder without the need for data labels such as trajectory-instruction pairs, object labels, etc.
Furthermore, since the agent can move in the interactive environments, we take advantage of this when designing the auxiliary tasks.
For example, different from other jigsaw-like self-supervised tasks that generate jigsaw puzzles by cropping images~\cite{doersch2015unsupervised,noroozi2016unsupervised} or by consecutive frames in video clips~\cite{ahsan2019video}, our proposed 3D jigsaw actively samples neighboring views, introducing more natural variations from the change of viewpoints.

\section{Method} \label{sec:method}

\begin{figure*}
\begin{subfigure}[t]{0.45\linewidth}
  \centering
  \includegraphics[width=\linewidth]{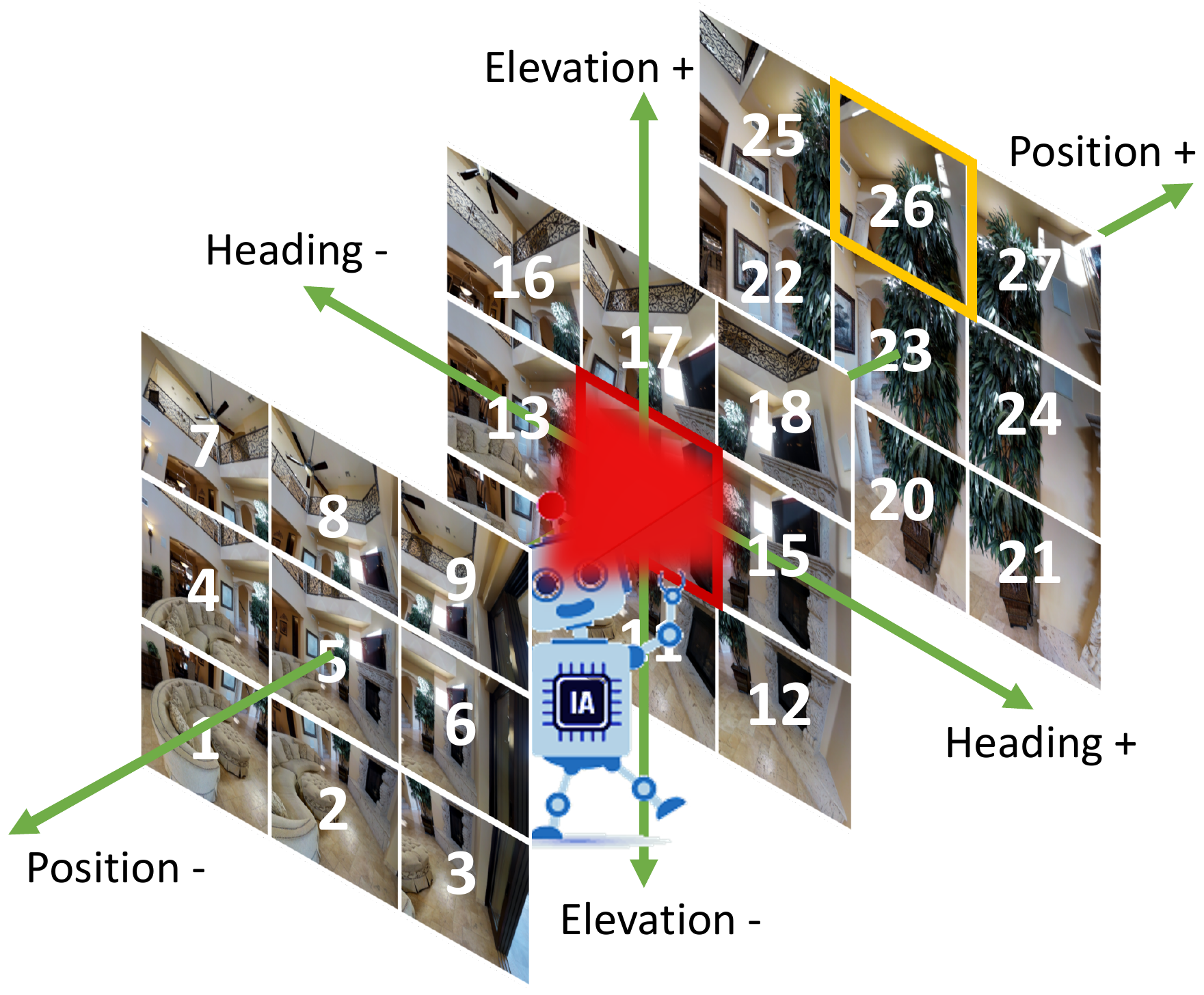}
  \caption{3D jigsaw}
  \label{fig:jigsaw}
\end{subfigure}
\begin{subfigure}[t]{0.3\linewidth}
  \centering
  \includegraphics[width=0.8\linewidth]{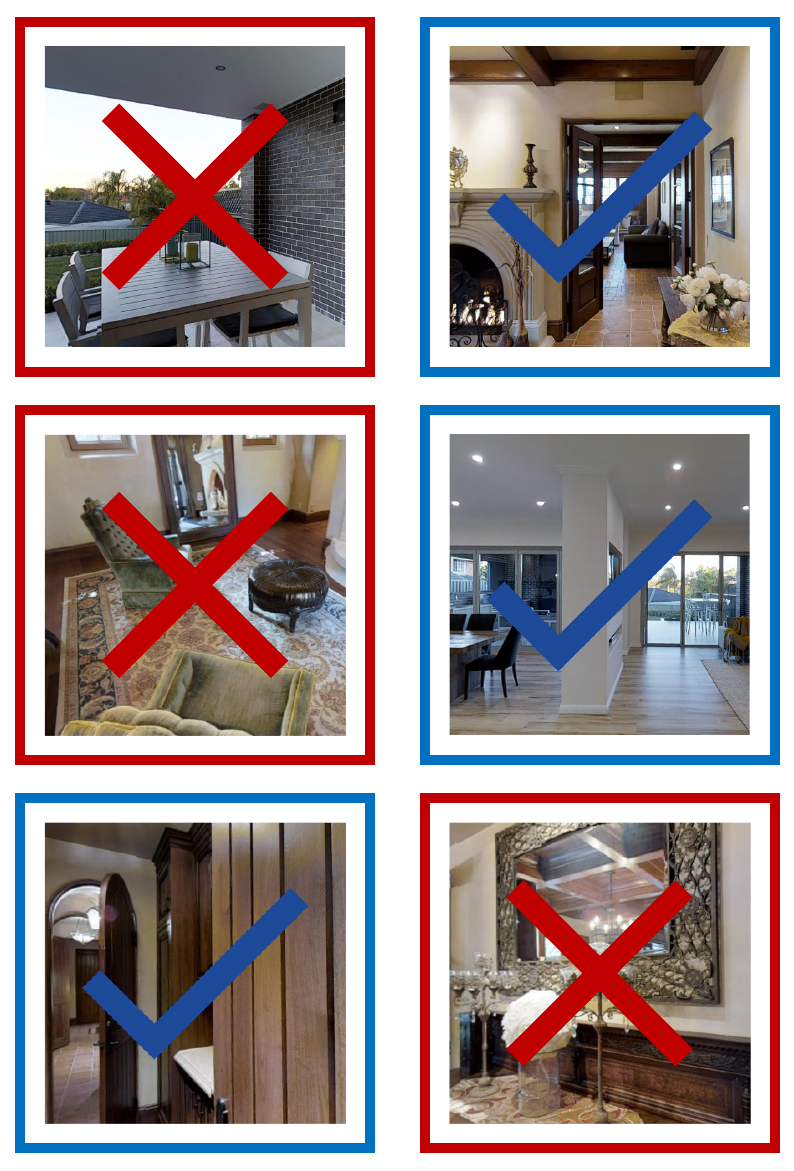}
  \caption{Traversability prediction}
  \label{fig:traversability}
\end{subfigure}
\begin{subfigure}[t]{0.2\linewidth}
  \centering
  \includegraphics[width=\linewidth]{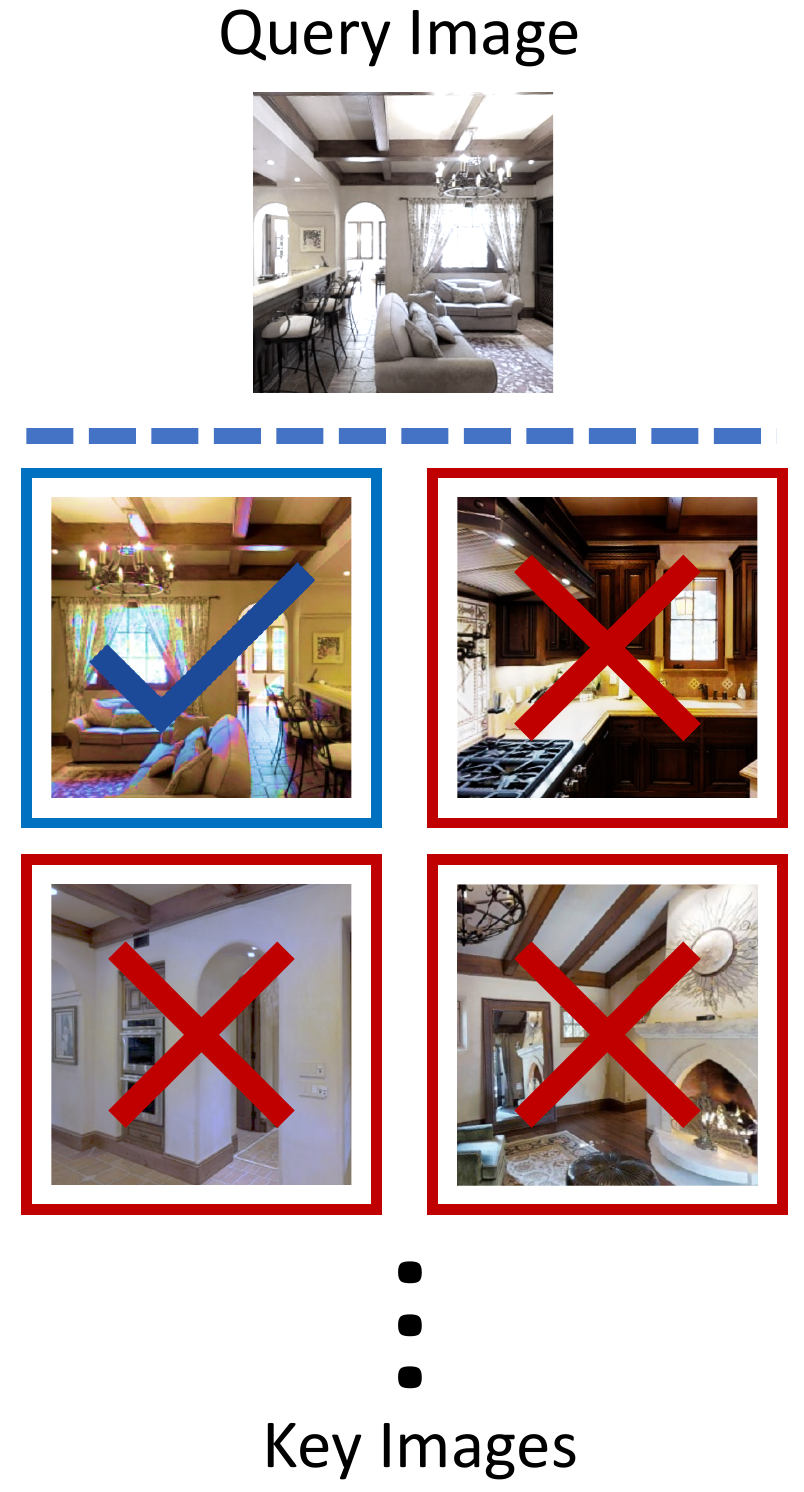}
  \caption{Instance Classification}
  \label{fig:instance-classification}
\end{subfigure}
\caption{
(Best viewed on the computer, in color and zoomed in.) We design three auxiliary tasks to encode structural information of the scenes, as well as the discriminative feature for object and scene classification crucial for the VLN task.
\textit{\textbf{(a)}} The auxiliary task of 3D jigsaw is to predict the relative pose between an anchor view (red box) and a query view (yellow).
The query view is sampled from the anchor view's neighboring views along the elevation, heading, and position dimensions.
\textit{\textbf{(b)}} The auxiliary task of traversability prediction is to predict whether a view contains any traversable direction.
The images in the blue box are labeled as \textit{True} (contain traversable directions), and the images in the red box are labeled as \textit{False} (do not contain traversable directions.)
\textit{\textbf{(c)}} The auxiliary task of instance classification is to identify a view's augmented copy from a pool of other image views.
In this example, the view in the blue box is the corresponding augmented copy (positive pair), while the views in the red box are other image views (negative pairs).
}
\label{fig:auxiliary-tasks}
\end{figure*}

In existing VLN methods, the ImageNet pre-trained features may be suboptimal due to the data distribution shift between ImageNet and the navigation environments, and the difference between the pre-training classification task and the target VLN task.
As explained in Section~\ref{sec:introduction}, the na\"ive solution of fine-tuning the image encoder in the target environments is inapplicable due to the lack of labeled images.
Furthermore, it is also computationally prohibitive to jointly train the image encoder with the agent on the target VLN task.
Therefore, we seek to design auxiliary tasks that can improve the image encoder but rely only on data available in the navigation environments.

\subsection{Problem Setting}
In VLN, a navigation agent is given training data in the form of trajectory-instruction pairs in different indoor environments.
In the training environments, the agent is also allowed to access data such as RGB image views, a view's neighboring views, and traversable directions contained in a view.
In this paper, for fair comparison with other works, the proposed auxiliary tasks are trained with only these information to make sure that the performance gain is \textit{not} coming from additional training signals (\textit{e.g.,} semantic segmentation map, room type).

\subsection{Auxiliary Tasks}

With the data collected from the environments, we aim to design auxiliary tasks that help the image encoder encode visual information that is crucial for the target VLN task.
To find out what are important features, we start by observing the following instruction example:
\textit{``\textbf{Exit} the \textbf{screening room}, make a \textbf{right}, go \textbf{straight into} the \textbf{room} with the \textbf{globe} and stop.''}
As highlighted above, to correctly follow the instruction, the agent needs to encode the following information from its image encoder: (1) structural information of the scene (exit, right, straight into), and (2) discriminative information for scenes and objects (screening room, room, globe) in the visual representation.
Therefore, we design three auxiliary tasks shown in Figure~\ref{fig:auxiliary-tasks}: (1) \textit{3D jigsaw}, (2) \textit{traversability prediction}, and (3) \textit{instance classification} to encode these crucial information for VLN.

\subsubsection{3D Jigsaw}
In order to follow the instructions correctly to reach the target location, the agent has to interpret instructions such as "turn \textbf{left}", "turn \textbf{right} when you see the sofa on your \textbf{left}", "stop \textbf{in front of} the TV" from the visual representation.
Therefore, we propose the auxiliary task of 3D jigsaw to encode structural information of the scene by predicting the relative poses (position, heading, and elevation) of two views.
As shown in Figure~\ref{fig:jigsaw}, given an \textit{anchor view} $x_a$ in the red box, a \textit{query view} $x_q$ in the yellow box is sampled from the neighboring views around the anchor view.
Neighboring views are views within the $[-1, 0, +1]$ range of discretized headings, elevations, and positions, forming a 3D jigsaw with 27 views ($3\times3\times3$).
The label of the neighboring views (jigsaw labels) can be uniquely determined by their relative poses to the anchor view (the ``numbers'' overlaid on the neighboring views in Figure~\ref{fig:jigsaw}).
If the sampled anchor view is looking up, neighboring views of \{7-9, 16-18, 25-27\} in Figure~\ref{fig:jigsaw} would be unavailable due to the way views are discretized (similarly for the case of looking down.)
On the other hand, if the sampled anchor view does not contain any traversable directions, neighboring views of \{19-27\} would be unavailable because the agent cannot go one step forward.
Nevertheless, the auxiliary task and jigsaw labels can still be constructed in a similar way.
Those unavailable neighboring views are simply removed.

Conditioned on the anchor view, the 3D jigsaw task is formulated as a 27-class classification problem.
The prediction $p_{jig}$ is computed by:
\begin{equation}
    p_{jig} = \text{softmax}(\phi_{jig}([f_{enc}(x_{a}),f_{enc}(x_{q})])),
\end{equation}
where $f_{enc}$ is the image encoder shared with other auxiliary tasks, $\phi_{jig}$ is a multi-layer perceptron specific for 3D jigsaw, and $[\cdot, \cdot]$ is a concatenation operation along the feature dimension.
The loss is simply a cross-entropy loss:
\begin{equation}
    \mathcal{L}_{jig} = - \frac{1}{N} \sum_{i}^{N} y_{i,jig} \log p_{i,jig},
\end{equation}
where $y_{i,jig}$ and $p_{i,jig}$ are the jigsaw label and prediction for the $i$-th training example, respectively, and the loss is averaged over a mini-batch of $N$ examples. 

\subsubsection{Traversability Prediction}
In order to encode the layout (structure) and navigation information of the scene and environments, we propose an auxiliary task of traversability prediction shown in Figure~\ref{fig:traversability}.
The image encoder classifies a given image view as \textit{true} when the view contains traversable directions, otherwise classifies it as \textit{false}.
Following the practice in Matterport3D (MP3D) simulator~\cite{anderson2018vision,Matterport3D}, a traversable direction is contained within the current view if a discretized traversable location is within the horizontal field of the current view and within 5 meters Euclidean distance of the current location.
This information is acquired by building and parsing the navigation graph of the environments and is provided in the MP3D simulator as well as many other VLN simulators and datasets~\cite{hermann2020learning,thomason2020vision,vasudevan2021talk2nav,chen2019touchdown}.

The traversability prediction task is formulated as a binary classification problem. The prediction $p_{nav}$ is computed as:
\begin{equation}
    p_{trav} = \sigma(\phi_{trav}(f_{enc}(x))),
\end{equation}
where $f_{enc}$ is the image encoder shared with other auxiliary tasks, $\phi_{trav}$ is a multi-layer perceptron specific for traversability prediction, and $\sigma$ is the sigmoid activation function.
The loss is simply a binary cross-entropy loss:
\begin{equation}
\begin{aligned}
    \mathcal{L}_{trav} = - \frac{1}{N} \sum_{i}^{N}\ &y_{i,trav} \log p_{i,trav} + \\
    &(1 - y_{i,trav}) \log (1 - p_{i,trav}),
\end{aligned}
\end{equation}
where $y_{i,trav}$ and $p_{i,trav}$ are the traversability label and prediction for the $i$-th training example, respectively, and the loss is averaged over a mini-batch of $N$ examples.

\subsubsection{Instance Classification} \label{sec:contrastive-learning}
To follow instructions correctly, the agent has to encode scene information such as kitchen, bedroom, bathroom, and so on, as well as object information such as chair, sofa, TV, etc.
In computer vision, instance classification \cite{he2020momentum,chen2020simple} has achieved remarkable progress in representation learning.
It has been shown that the representation learned by instance classification transfers well to many downstream tasks, such as object classification, object detection, scene classification.
Therefore, we apply instance classification as an auxiliary task in the navigation environments to encode discriminative information for objects and scenes.

As shown in Figure~\ref{fig:instance-classification}, given an image view $x$, we generate a query image $x_q$ and a key image $x_k$ by applying image augmentations, such as color jittering, lighting adjustment, affine transform, etc, on the image view $x$.
Given the query image $x_q$, instance classification task is to identify the corresponding key image $x_k$ (positive sample) from a pool of other image views (negative samples).
Similar to MoCo~\cite{he2020momentum}, we use a memory bank to increase the number of negative samples by storing the encoded features of training samples from previous mini-batches.
We also use the current image encoder to encode $x_q$ and the moving-averaged image encoder to encode $x_k$.
The instance prediction $p_{ins}$ can be computed as:
\begin{equation}
    p_{ins} = \frac{\exp(\phi_{ins}(f_{enc}(x_q))\cdot \hat{\phi}_{ins}(\hat{f}_{enc}(x_k)/\tau))}{\sum_{i}\exp(\phi_{ins}(f_{enc}(x_q))\cdot m_i/\tau)},
\end{equation}
where $f_{enc}$ is the current image encoder shared with other auxiliary tasks, $\phi_{ins}$ is a multi-layer perceptron specific for instance classification, $\hat{f}_{enc}$ and $\hat{\phi}_{ins}$ are their moving-averaged version, $m_i$ is the $i$-th entry (negative samples) in the memory bank, and $\tau$ is a scaling factor (temperature).
The loss is simply a cross-entropy loss:
\begin{equation}
    \mathcal{L}_{ins} = - \frac{1}{N} \sum_{i}^{N} y_{i,ins} \log p_{i,ins},
\end{equation}
where $y_{i,ins}$ and $p_{i,ins}$ are the label for positive samples and instance prediction for the $i$-th training example, respectively, and the loss is averaged over a mini-batch of $N$ examples.

\begin{figure}
\centering
\includegraphics[width=1.0\linewidth]{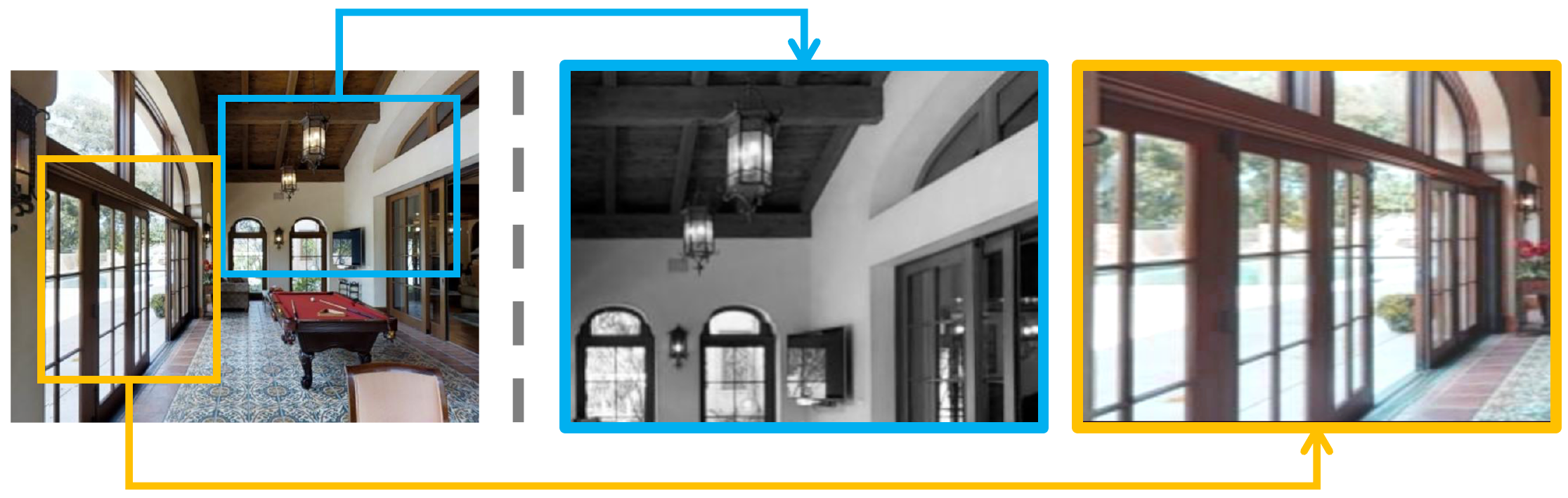}
\caption{
Applying aggressive resize crop augmentation that is effective on object-centric images may remove important visual clues in scene images with multiple objects.
In this example, it is ambiguous to classify the image in blue with the image in yellow as a positive pair.
}
\label{fig:instance-aug}
\end{figure}

To learn a good visual representation by instance classification, image augmentations play a crucial role~\cite{chen2020simple,chen2020improved}.
Good image augmentations depend on the downstream task, as well as the form of data on which instance classification is applied~\cite{tian2020makes}.
It has been shown that color jittering, Gaussian blur, horizontal flip, and resize crop are particularly useful for learning on datasets with object-centric images such as ImageNet~\cite{chen2020simple}.
In navigation environments, however, image views contain multiple objects in a scene.
Aggressive resize crop (scale ranged between $[0.2, 1.0]$) may remove important information and lead to ambiguous situations as illustrated in Figure~\ref{fig:instance-aug}.
Hence, we use a weak resize crop (scale ranged between $[0.8, 1.0]$) with an affine transform in place of the aggressive resize crop.

\subsection{Training Procedure}\label{sec:training_procedure}
In this section, we explain how to train the image encoder efficiently and how to train the VLN agent with our SEA pre-trained features.

\subsubsection{Image Encoder}
We propose a training procedure to reuse the data in a mini-batch across three auxiliary tasks.
Without data reuse, we need two training samples for 3D jigsaw, one for traversability, and one for instance classification, which sum up to four training samples.
This will be expensive in terms of computation and memory usage especially for loading and training on the high-resolution images in VLN.

Given the \textit{i}-th image $x_i$ in a mini-batch, in 3D jigsaw we use $x_i$ as the anchor view and sample a query view $x_{i,q}$ from $x_i$'s neighboring views.
$x_i$ is reused for traversability prediction.
In instance classification, we again reuse $x_i$ as the query image view, and its augmented copy as the key image view $x_{i,k}$.
Two images are sampled in total.
To further reduce computation, only $x_i$ is fed into the current model for backpropagation.
All the other image views including $x_{i,q}$ from 3D jigsaw and $x_{i,k}$ from instance classification are fed into the moving-averaged image encoder inspired by MoCo~\cite{he2020momentum}.
In this way, we can drastically save computation since images used for different auxiliary tasks are shared.
Furthermore, the features computed by the moving-averaged image encoder reduce memory usage by not constructing the computation graph and reduce computation by not performing backpropagation.

Finally, the image encoder is optimized by the sum of losses from the three auxiliary tasks:
\begin{equation}
    \mathcal{L} = \lambda_{jig} \mathcal{L}_{jig} + \lambda_{trav} \mathcal{L}_{trav} + \lambda_{ins} \mathcal{L}_{ins}
\end{equation}
We empirically set $\lambda_{jig}=\lambda_{trav}=\lambda_{ins}=1$ without further hyper-parameter tuning.

\subsubsection{Agent}
After pre-training the image encoder with the proposed auxiliary tasks, we pre-compute the features for each discretized view in the training environments following the convention in \cite{anderson2018vision} and all other VLN works.
The VLN agent then uses our pre-computed features of each view in place of the ImageNet pre-trained features for training.
By decoupling the training of the image encoder from the VLN agent, other VLN methods can benefit from our improved visual representation with minimal modification.

\section{Experiments}

\subsection{Dataset}
In this paper, we propose and validate our method on the Matterport3D (MP3D) simulator~\cite{anderson2018vision,Matterport3D} and Room-to-Room (R2R) dataset~\cite{anderson2018vision}
, but the method is applicable in general navigation  settings~\cite{hermann2020learning,thomason2020vision,vasudevan2021talk2nav,chen2019touchdown} where neighboring image views and traversability information are available.

\textbf{Dataset for pre-training the image encoder.}
To pre-train the image encoder, we collect data available in the environments of the MP3D simulator such as RGB image views, a view's neighboring views, traversable directions within a view, etc.
Following Anderson et al.~\cite{anderson2018vision}, at each location, the views are discretized at $30^\circ$ interval in the range of $[0^\circ, 330^\circ]$ for heading, and $[-30^\circ, 30^\circ]$ for elevation, resulting in 36 views at each location.
Following the environment splits in \cite{anderson2018vision}, the pre-training dataset is composed of around 275k discretized image views in the Train environments, 34k in the Val-Unseen, and 71k in the Test-Unseen.
The image encoder is only pre-trained on data from the Train environments.

\textbf{Dataset for training the VLN agent.}
We use the Room-to-Room (R2R) dataset~\cite{anderson2018vision} which contains 7,189 training data in the form of human instruction and trajectory pairs.
Each trajectory is paired with three instructions.
The whole dataset is divided into four sets: Train, Val-Seen, Val-Unseen, and Test-Unseen.
The Val-Seen environments are the same as the Train environments but with different navigation instructions. 
On the other hand, the Val-Unseen and Test-Unseen environments are different with the Train environments and also with different navigation instructions.

\subsection{Evaluation}
The effectiveness of the pre-trained features is evaluated by the performance of the agent on the target VLN task.
Since the training of the image encoder and the agent are decoupled, the performance improvement of the agent can be solely attributed to the improvement of the image representation.
The agent is evaluated on both seen (Val-Seen), and unseen (Val-Unseen and Test-Unseen) environments.
Even though benchmarking on seen environments (Val-Seen) has been conducted, the primary goal of the VLN agents is to learn to generalize well on unseen environments (Val-Unseen and Test-Unseen).
Following \cite{anderson2018vision,anderson2018evaluation} and other VLN methods, the agent is evaluated with the following metrics: (1) \textbf{TL}: average trajectory length, (2) \textbf{NE}: navigation error defined as the average shortest path distance between the agent's final location and the target location, (3) \textbf{SR}: success rate defined as the percentage of agent's final location within three meters from the target location, and (4) \textbf{SPL}: SR weighted by path length that penalize SR by TL.

\begin{table*}
\centering
\renewcommand{\arraystretch}{1.2}
\resizebox{\textwidth}{!}{
\begin{tabular}{@{\extracolsep{4pt}}lcccc|cccc|cccc@{}}
\toprule
& \multicolumn{4}{c}{\textbf{Val-Seen}} & \multicolumn{4}{c}{\textbf{Val-Unseen}} & \multicolumn{4}{c}{\textbf{Test-Unseen}}\\
\cline{2-5} \cline{6-9} \cline{10-13}
Method & TL & NE$\downarrow$ & SR$\uparrow$ & \multicolumn{1}{c}{SPL$\uparrow$} & TL & NE$\downarrow$ & SR$\uparrow$ & \multicolumn{1}{c}{SPL$\uparrow$} & TL & NE$\downarrow$ & SR$\uparrow$ & SPL$\uparrow$ \\
\midrule
RCM~\cite{wang2019reinforced} & 10.65 & 3.53 & 0.67 & - & 11.46 & 6.09 & 0.43 & - & 11.97 & 6.12 & 0.43 & 0.38 \\
Self-Monitoring~\cite{ma2019selfmonitoring} & - & \textbf{3.22} & 0.67 & 0.58 & - & 5.52 & 0.45 & 0.32 & 18.04 & 5.67 & 0.48 & 0.35 \\
Regretful Agent~\cite{ma2019regretful} & - & 3.23 & 0.69 & 0.63 & - & 5.32 & 0.50 & 0.41 & 13.69 & 5.69 & 0.48 & 0.40 \\

PREVALENT~\cite{hao2020towards} & 10.32 & 3.67 & 0.69 & 0.65 & 10.19 & 4.71 & \textbf{0.58} & \textbf{0.53} & 10.51 & 5.30 & 0.54 & 0.51 \\
Relationship Graph~\cite{hong2020language} & 10.13 & 3.47 & 0.67 & 0.65 & 9.99 & 4.73 & 0.57 & \textbf{0.53} & 10.29 & 4.75 & 0.55 & 0.52 \\
Speaker-Follower~\cite{fried2018speaker} & - & 3.36 & 0.66 & - & - & 6.62 & 0.35 & - & 14.82 & 6.62 & 0.35 & 0.28 \\
Env-Dropout~\cite{tan2019learning} & 11.00 & 3.99 & 0.62 & 0.59 & 10.70 & 5.22 & 0.52 & 0.48 & 11.66 & 5.23 & 0.51 & 0.47 \\
AuxRN~\cite{zhu2020vision} & - & 3.33 & \textbf{0.70} & \textbf{0.67} & - & 5.28 & 0.55 & 0.50 & - & 5.15 & 0.55 & 0.51 \\
\hline
Speaker-Follower + SEA (ours) & 12.80 & 3.68 & 0.64 & 0.56 & 13.61 & 5.16 & 0.51 (+16\%) & 0.42 & 14.07 & 5.42 & 0.47 (+12\%) & 0.40 (+12\%) \\
Env-Dropout + SEA (ours) & 10.31 & 3.44 & 0.69 & 0.66 & 9.88 & 4.76 & 0.56 (+4\%) & 0.52 (+4\%) & 10.18 & 4.89 & 0.56 (+5\%) & 0.53 (+6\%) \\
AuxRN + SEA (ours) & 10.28 & 3.43 & 0.68 & 0.65 & 9.80 & \textbf{4.55} & 0.57 (+2\%) & \textbf{0.53} (+3\%) & 10.31 & \textbf{4.71} & \textbf{0.59} (+4\%) & \textbf{0.55} (+4\%) \\
\bottomrule
\end{tabular}
}
\caption{
Comparison to other classical VLN methods under the single-run setting, where the image encoder and the agent have no access to unseen environments (Val-Unseen and Test-Unseen) during training.
The Speaker-Follower~\cite{fried2018speaker}, Env-Dropout~\cite{tan2019learning}, and AuxRN~\cite{zhu2020vision} methods combined with our SEA features achieve significant performance improvement on both Val-Unseen and Test-Unseen sets.
}
\label{table:single-run}
\end{table*}

\subsection{Main Results}
We first show that our pre-trained features are superior to the ImageNet pre-trained features, and can boost navigation agents' performance by simply training with our SEA pre-trained features in place of the ImageNet pre-trained features. 
The agent is evaluated under the single-run setting, where only data from the training environments are available for training both the agent and the image encoder.
No extra information is included in comparison with other VLN methods since the image encoder is also pre-trained only on data from the training environments.
The single-run setting tests the generalization performance of both the agent and the visual representation to new held-out environments.
For the VLN agents, we select Speaker-Follower~\cite{fried2018speaker}, Env-Dropout~\cite{tan2019learning}, and AuxRN~\cite{zhu2020vision}, and replace the ImageNet pre-trained features with our SEA pre-trained features.
We use the released code from these VLN methods and train the agent with our SEA pre-trained features without any hyper-parameter tuning for the agent.

In Table~\ref{table:single-run}, with our SEA pre-trained features, all three agents achieve consistent improvement in Val-Unseen and Test-Unseen.
Notably, in Test-Unseen, the most important part of the evaluation since it tests \textit{generalization} performance to new held-out environments, our SEA pre-trained features achieve 12\% absolute improvement in both SR and SPL for Speaker-Follower, and 4\% for the already strong AuxRN agent.
The improvement we obtain can be solely attributed to our pre-trained features, not to the improvement of the agents as we didn't tune the agent at all.
We anticipate higher performance is possible with tuning.
These results also highlight the importance of visual representation that has long been ignored in the VLN task.
Furthermore, since the improvement of the visual representation is orthogonal to the improvement of the agent, other VLN agents and follow-on works can also benefit from and build on our SEA pre-trained features.
We will release the pre-training dataset, source code, and SEA pre-trained features to facilitate future research in VLN.

\subsection{Analysis}

\begin{table}
\centering
\renewcommand{\arraystretch}{1.2}
\resizebox{1.0\columnwidth}{!}{
\begin{tabular}{@{\extracolsep{4pt}}lccc@{}}
\toprule
& 3D Jigsaw & Traversability & Instance Classif. \\
\midrule
Initial accuracy & 5.19 & 64.12 & 0.62 \\
Final accuracy & 50.83 & 89.85 & 99.86 \\
\bottomrule
\end{tabular}
}
\caption{
The classification accuracy (in percentage) of each auxiliary task at the beginning and the end of training.
}
\label{table:analysis:auxiliary}
\end{table}

\textbf{Does the image encoder indeed learn to perform well on the auxiliary tasks?}
Since the agent's improvement is coming from the improved visual representation, which is coming from the training on the three proposed auxiliary tasks, we first verify that the proposed auxiliary tasks are learnable and the image encoder indeed learns to perform well on the tasks.
We report the performance (accuracy in percentage) on a held-out validation set at the beginning of training and at the end of training for each auxiliary task.
Note that the image views in the held-out validation set are collected from Val-Unseen environments different from the training environments.

The results are shown in Table~\ref{table:analysis:auxiliary}.
The image encoder indeed learns to do well on all the auxiliary tasks.
The accuracy numbers should not be compared across different auxiliary tasks as they vary in difficulty both because of the number of ``categories'' but also due to intrinsic difficulty (\textit{e.g.,} jigsaw is known to be harder~\cite{doersch2015unsupervised}.)

\begin{table*}
\centering
\renewcommand{\arraystretch}{1.2}
\resizebox{1.0\textwidth}{!}{
\begin{tabular}{@{\extracolsep{4pt}}cccc|cccc@{}}
\toprule
& \multicolumn{3}{c}{\textbf{Conditions}} & \multicolumn{4}{c}{\textbf{Downstream Tasks}} \\
\cline{2-4} \cline{5-8}
& & & \multicolumn{1}{c}{} & Semantic Segmentation & Normal Estimation & Object Classif. & Scene Classif. \\
& 3D Jigsaw & Traversability Pred. & \multicolumn{1}{c}{Instance Classif.} & (mAP) $\uparrow$ & (RMSE) $\downarrow$ & (mAP) $\uparrow$ & (accuracy) $\uparrow$ \\
\midrule
ImageNet & - & - & - & 29.40 & 0.585 & 36.63 & \textbf{71.48} \\
\#2 (all) & \checkmark & \checkmark & \checkmark & \textbf{40.27} & \textbf{0.523} & \textbf{36.86} & 69.88 \\
\#3 & \checkmark & & & 30.32 (-25\%) & 0.557 (+7\%) & 27.07 (-27\%) & 61.42 (-12\%) \\
\#4 & & \checkmark & & 23.69 (-41\%) & 0.568 (+8\%) & 27.32 (-26\%) & 58.17 (-17\%) \\
\#5 & & & \checkmark & 35.76 (-11\%) & 0.545 (+4\%) & 33.72 (-9\%) & 69.64 (-0\%) \\
\#6 & \checkmark & \checkmark & & 34.12 (-15\%) & 0.546 (+4\%) & 29.2 (-21\%) & 63.67 (-9\%) \\
\#7 & \checkmark & & \checkmark & 37.46 (-7\%) & 0.533 (+2\%) & 34.21 (-7\%) & 70.12 (0\%) \\
\#8 & & \checkmark & \checkmark & 37.69 (-6\%) & 0.540 (-3\%) & 32.56 (-12\%) & 68.99 (-1\%) \\
\bottomrule
\end{tabular}
}
\caption{
The analysis of what information is encoded by which auxiliary task.
The number in the parenthesis at row \#3 - \#8 represents the relative difference with respect to the full model with all auxiliary tasks combined (row \#2).
}
\label{table:analysis:feature}
\end{table*}

\begin{table*}
\centering
\renewcommand{\arraystretch}{1.2}
\resizebox{1.0\textwidth}{!}{
\begin{tabular}{@{\extracolsep{4pt}}cccc|cccc|cccc@{}}
\toprule
& \multicolumn{3}{c}{\textbf{Conditions}} &  \multicolumn{4}{c}{\textbf{Val-Seen}} & \multicolumn{4}{c}{\textbf{Val-Unseen}} \\
\cline{2-4} \cline{5-8} \cline{9-12}
& 3D Jigsaw & Traversability Pred. & \multicolumn{1}{c}{Instance Classif.} & TL & NE$\downarrow$ & SR$\uparrow$ & \multicolumn{1}{c}{SPL$\uparrow$} & TL & NE$\downarrow$ & SR$\uparrow$ & SPL$\uparrow$ \\
\midrule
\#1 (all) & \checkmark & \checkmark & \checkmark & 10.31 & \textbf{3.44} & \textbf{0.69} & \textbf{0.66} & \textbf{9.88} & \textbf{4.76} & \textbf{0.56} & \textbf{0.52} \\
\#2 & \checkmark & & & 10.04 & 4.67 & 0.57 (-12\%) & 0.55 (-11\%) & 9.58 & 5.22 & 0.52 (-4\%) & 0.49 (-3\%)\\
\#3 & & \checkmark & & 10.15 & 5.89 & 0.47 (-22\%) & 0.44 (-21\%) & 9.52 & 5.93 & 0.47 (-9\%) & 0.43 (-8\%) \\
\#4 & & & \checkmark & 10.46 & 3.74 & 0.64 (-5\%) & 0.62 (-4\%) & 9.93 & 5.37 & 0.53 (-3\%) & 0.49 (-3\%) \\
\#5 & \checkmark & \checkmark & & 10.21 & 4.33 & 0.62 (-7\%) & 0.58 (-7\%) & 9.87 & 5.08 & 0.53 (-3\%) & 0.49 (-3\%) \\
\#6 & \checkmark & & \checkmark & 10.41 & 3.93 & 0.65 (-4\%) & 0.62 (-4\%) & 9.65 & 4.83 & 0.55 (-1\%) & 0.51 (-1\%) \\
\#7 & & \checkmark & \checkmark & 10.36 & 3.82 & 0.66 (-3\%) & 0.63 (-3\%) & 10.21 & 5.33 & 0.53 (-3\%) & 0.49 (-3\%) \\
\bottomrule
\end{tabular}
}
\caption{
The correlation between agent's navigation performance and the features pre-trained with different sets of auxiliary tasks.
The number in the parenthesis at row \#2 - \#7 represents the absolute difference with respect to the full model (row \#1.)
}
\label{table:analysis:agent}
\end{table*}

\textbf{What information is encoded by training on the auxiliary tasks?}
Now that we know the learned image encoder does well on the auxiliary tasks, we further analyze what information is encoded in the features.
Therefore, we conduct ablation studies on the pre-trained image encoder.
Specifically, we first train the image encoder with different combinations of auxiliary tasks.
We then append a light-weight head to the image encoder and fine-tune only the head (with the image encoder frozen) to downstream tasks: (1) semantic segmentation, (2) normal estimation, (3) multi-label object classification, and (4) scene classification.
The training data are taken from a subset of the Taskonomy~\cite{taskonomy2018} dataset.
Semantic segmentation and normal estimation require structural information of the scene, while multi-label object classification and scene classification require discriminative information of objects and classes.

The results are shown in Table~\ref{table:analysis:feature}.
We first compare the full model (\#2) and the ImageNet features.
The full model performs significantly better in semantic segmentation and normal estimation while maintaining slightly better or comparable performance in multi-label object classification and scene classification.
This explains why the agent trained with our SEA pre-trained features performs much better in the target VLN task: \textit{our SEA pre-trained features successfully encode more structural information of the scenes}, which are crucial for performing a navigation task in addition to the discriminative information of objects and scenes.
For example, agents frequently make decisions based on instructions like: "turn right when you see the sofa on your left," which requires the understanding of the structure of the scene to successfully follow.

Next, we observe that instance classification (\#5,7,8) is the most effective auxiliary task across all downstream tasks.
Even though 3D jigsaw and traversability do not perform well by themselves (\#3,4), when combined with instance classification (\#7,8 compared with \#5), they are beneficial for encoding structural information of the scenes as they provide substantial gains in semantic segmentation and normal estimation.
Furthermore, 3D jigsaw also provides marginal gains in multi-label object classification and scene classification (\#7 compared with \#5).

\textbf{How does agent's performance correlate with each auxiliary task?}
Now that we know which auxiliary task helps encode what kind of information, we would like to assess whether this encoded information is truly beneficial to the agent's final navigation performance.
Therefore, we conduct ablation studies under the single run setting with the Env-Drop agent~\cite{tan2019learning}.
Specifically, we first train the image encoder with different combinations of auxiliary tasks, use the pre-trained image encoder to generate pre-computed features, and train the agent with the pre-computed features.

The results on Val-Seen and Val-Unseen are shown in Table~\ref{table:analysis:agent}.
The agent's performance drops when any of the auxiliary tasks are removed while training the image encoder.
Similar to what we've found previously, instance classification (\#4,6,7) is the most effective auxiliary task among the three, while 3D jigsaw and traversability are also beneficial as they further improve the performance when combined with instance classification.
For example, on top of instance classification, 3D jigsaw helps the agent perform even better on Val-Unseen (\#6 compared with \#4).
Traversability prediction does not help much on Val-Unseen but is beneficial on Val-Seen (\#7 compared with \#4.)
This could be explained by navigation graphs (traversability) of the environments providing a strong prior of the environment layout~\cite{krantz2020beyond}.
Thus, the auxiliary task of traversability prediction learns to encode the prior, adapting to the environments.

\section{Conclusion}
We propose structure-encoding auxiliary tasks (SEA) to improve the visual representation, long ignored in VLN.
Three auxiliary tasks, 3D jigsaw, traversability prediction, and instance classification, are proposed and customized to pre-train the image encoder on data gathered in the navigation environments.
3D jigsaw and instance classification help better encode both structural information of scenes and discriminative information of objects and scenes, while traversability prediction helps better encode structural information and adapt the visual representation to the target navigation environments.
The VLN agents combined with our SEA pre-trained features (without tuning) achieve 12\% SR improvement for Speaker-Follower, 5\% for Env-Dropout, and 4\% for AuxRN in test-unseen under the single-run setting.
The contributions of proposed auxiliary tasks and SEA pre-trained features are orthogonal to other VLN works, and we will release the collected dataset, source code, and pre-trained features to facilitate further research in visual representations for VLN.

\clearpage
\begin{table*}[bp]
\centering
\renewcommand{\arraystretch}{1.2}
\resizebox{0.95\textwidth}{!}{
\begin{tabular}{@{\extracolsep{4pt}}lcccc|cccc@{}}
\toprule
& \multicolumn{4}{c}{\textbf{Pre-training Environments}} & \multicolumn{4}{c}{\textbf{Val Unseen}} \\
\cline{2-5} \cline{6-9}
Method & Train & Val-Unseen & Test-Unseen & \multicolumn{1}{c}{\#data} & TL & NE$\downarrow$ & SR$\uparrow$ & SPL$\uparrow$ \\
\midrule
Env-Drop baseline (ImageNet) \cite{tan2019learning} & - & - & - & - & 10.70 & 5.22 & 0.52 & 0.48 \\
Env-Drop + SEA (ours) & \checkmark & & & 275k & 10.024 & 4.678 & 0.555 & 0.518 \\
& \checkmark & \checkmark & & 309k (+12\%) & 9.959 & 4.705 & 0.566 & 0.529 \\
& \checkmark & \checkmark & \checkmark & 380k (+38\%) & 9.855 & 4.656 & \textbf{0.573} & \textbf{0.535} \\
\hline
AuxRN baseline (ImageNet) \cite{zhu2020vision} & - & - & - & - & - & 5.28 & 0.55 & 0.50 \\
AuxRN + SEA (ours) & \checkmark & & & 275k & 9.796 & 4.554 & 0.566 & 0.530 \\
& \checkmark & \checkmark & & 309k (+12\%) & 9.923 & 4.700 & 0.569 & 0.532 \\
& \checkmark & \checkmark & \checkmark & 380k (+38\%) & 9.937 & 4.566 & \textbf{0.572} & \textbf{0.534} \\
\bottomrule
\end{tabular}
}
\caption{
The performance of the Env-Dropout and AuxRN agents when trained with SEA features pre-trained with more data.
With more data available for pre-training the image encoder, both the Env-Dropout and AuxRN agents achieve better performance in the target navigation task on the Val-Unseen environments.
}
\label{supp:table:more-data}
\end{table*}

\section*{Appendix}
\appendix

\section{Does more pre-training data help?}
We are interested in whether an image encoder trained with our proposed SEA tasks can learn a representation that better transfers to the downstream navigation task if more pre-training data are available.
In self-supervised learning, it has been shown~\cite{yalniz2019billion,chen2020big,xie2020self} that image encoders pre-trained on web-scale datasets achieve better performance on downstream tasks.
Similarly, Majumdar et al.~\cite{majumdar2020improving} improves an agent's performance by pre-training a joint image-text representation on additional image captioning datasets.
If this is also true for our SEA pre-training, we could pre-train the image encoder on simulated environments or on real-world environments, as pre-training on the SEA tasks does not require the expensive trajectory-instruction pairs.
This can potentially improve the performance on the downstream navigation task.
To verify this, we pre-train the image encoder with the Val-Unseen and Test-Unseen environments, then train the Env-Dropout~\cite{tan2019learning} and AuxRN~\cite{zhu2020vision} agents with these SEA pre-trained features, and finally evaluate the agents' performance on the target navigation task on the R2R dataset.

The results are shown in Table~\ref{supp:table:more-data}.
We can see that the more data we have for pre-training the image encoder, the better performance the navigation agents could achieve.
For example, the Env-Dropout agent achieves 1.1\% SR improvement with 12\% more data (Train + Val-Unseen), and 1.8\% SR improvement with 38\% more data (Train + Val-Unseen + Test-Unseen).
This opens the possibility of pre-training on other simulated environments (\textit{e.g.}, Habitat~\cite{habitat19iccv} or AI2-THOR~\cite{kolve2017ai2}) or even on real-world environments for improving the navigation agent.
Furthermore, by pre-training the image encoder on the real-world environments, the image-encoder could potentially bridge the performance gap between the agents trained on simulated environments and the agents deployed in the real world.
We leave these interesting directions as future work.

\begin{table}
\centering
\renewcommand{\arraystretch}{1.2}
\resizebox{0.65\linewidth}{!}{
\begin{tabular}{@{\extracolsep{4pt}}l|c@{}}
\toprule
image resolution & $640(\text{w})\times480(\text{h})$ \\
vertical FOV & $60^\circ$ \\
orientation range & $[0^\circ, 30^\circ, ..., 330^\circ]$ \\
elevation range & $[-30^\circ, 0^\circ, 30^\circ]$ \\
\bottomrule
\end{tabular}
}
\caption{
Implementation details for collecting pre-training data.
}
\label{supp:table:image-format}
\end{table}

\begin{table}
\centering
\renewcommand{\arraystretch}{1.2}
\resizebox{0.95\linewidth}{!}{
\begin{tabular}{@{\extracolsep{4pt}}l|c@{}}
\toprule
batch size & 100 \\
training iterations & 50k \\
backbone & ResNet-50 \\
initialization & ImageNet \\
momentum image encoder & 0.999 \\
image normalization & mean [0.485, 0.456, 0.406] \\
 & std [0.229, 0.224, 0.225] \\
\hline
optimizer & SGD \\
momentum & 0.95 \\
Nesterov momentum & True \\
weight decay & 1e-4 \\
initial learning rate & 0.03 \\
learning rate scheduler & cosine \\
\hline
memory bank size & 32,768 \\
temperature $\tau$ & 0.07 \\
\bottomrule
\end{tabular}
}
\caption{Implementation details for the image encoder and the training procedure.}
\label{supp:table:image-encoder}
\end{table}

\section{Implementation Details}

We provide implementation details of how the data for pre-training the image encoder are collected in Tab.~\ref{supp:table:image-format}.
The implementation details for training the image encoder is in Tab.~\ref{supp:table:image-encoder}.
As for the VLN agent, please refer to the papers and released code of Speaker-Follower\footnote{\url{https://github.com/ronghanghu/speaker_follower}}~\cite{fried2018speaker}, Env-Dropout\footnote{\url{https://github.com/airsplay/R2R-EnvDrop}}~\cite{tan2019learning}, and AuxRN\footnote{\url{https://github.com/ZhuFengdaaa/MG-AuxRN}}~\cite{zhu2020vision}
as we do not tune any hyperparameters or training procedures of the VLN agents.

For the training of the image encoder, we reuse most of the hyperparameters from the released code of MoCov2~\cite{chen2020mocov2}.
We use a smaller batch size of 100 and training iterations of 50k so that the high-resolution images can fit into a machine with four NVIDIA 2080 Ti and the training can be done in about 24 hrs with automatic mixed precision (AMP) training.
To speed up training, we initialize the image encoder with ImageNet pre-trained weights.
Note that this does not include additional information compared to other VLN agents as they also use an ImageNet pre-trained image encoder.
We also half the size of the memory bank for contrastive learning as we use smaller batch size and the dataset contains much fewer training examples (see Tab.~\ref{supp:table:image-encoder}).
Lastly, to replace the data augmentation of strong resize crop (SimCLR~\cite{chen2020simple}) in Contrastive Learning, we use (1) random resize crop with scaling in the range of $[0.8, 1.0]$ and cropped output size of [430(h), 570(w)], and (2) random affine transform with rotation and shearing in the range of $[-10^\circ, 10^\circ]$.
Both (1) and (2) follow the definition in torchvision v1.6.

\section{Qualitative Results}

\begin{figure*}
\centering
\includegraphics[width=1\linewidth]{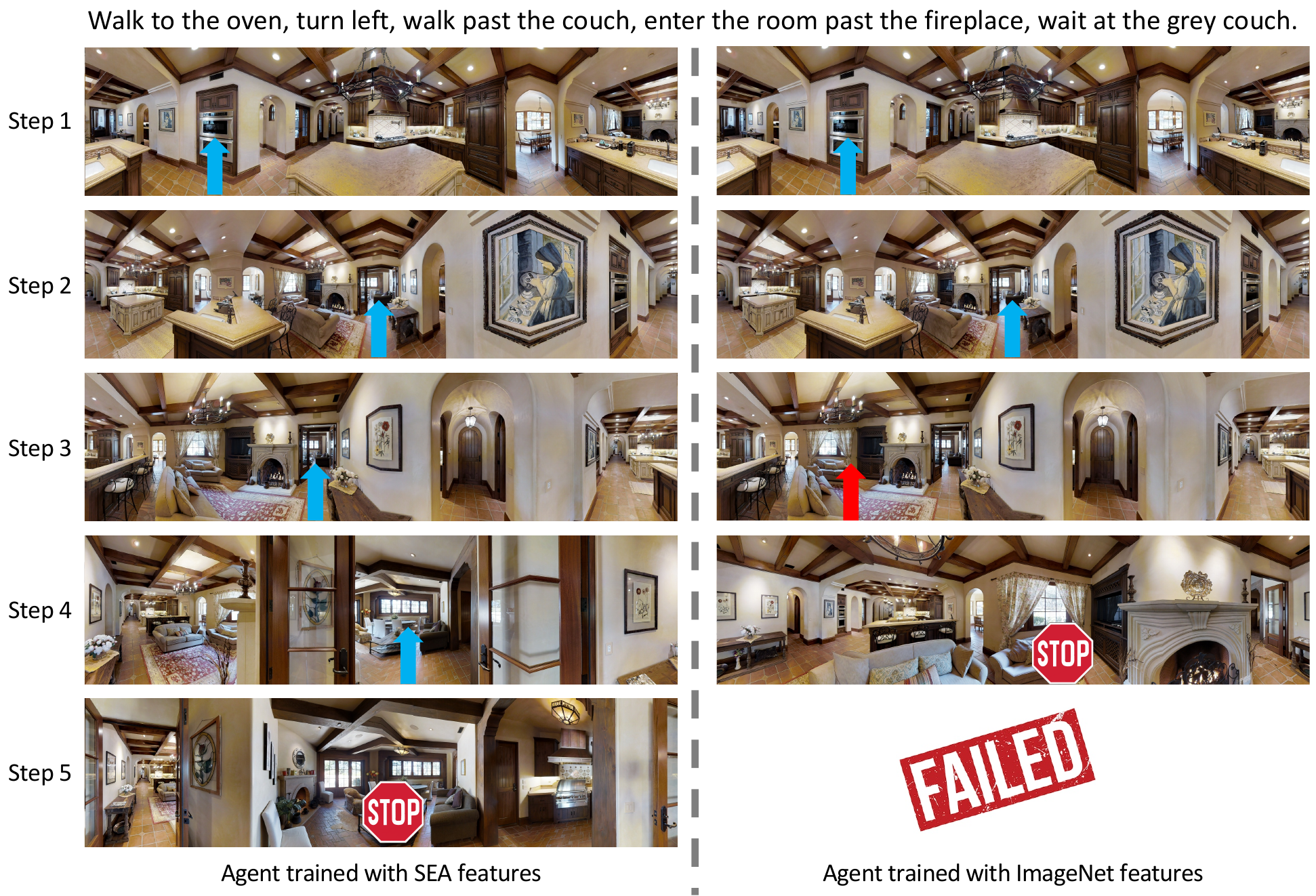}
\caption{
Step-by-step navigation results where the agent trained with SEA pre-trained features succeeds while the agent trained with ImageNet pre-trained features fails.
The agent trained with ImageNet pre-trained features fails at the step of ``\textit{enter the room past the fireplace}''.
}
\label{supp:fig:sea1im0:object}
\end{figure*}

\begin{figure*}
\centering
\includegraphics[width=1\linewidth]{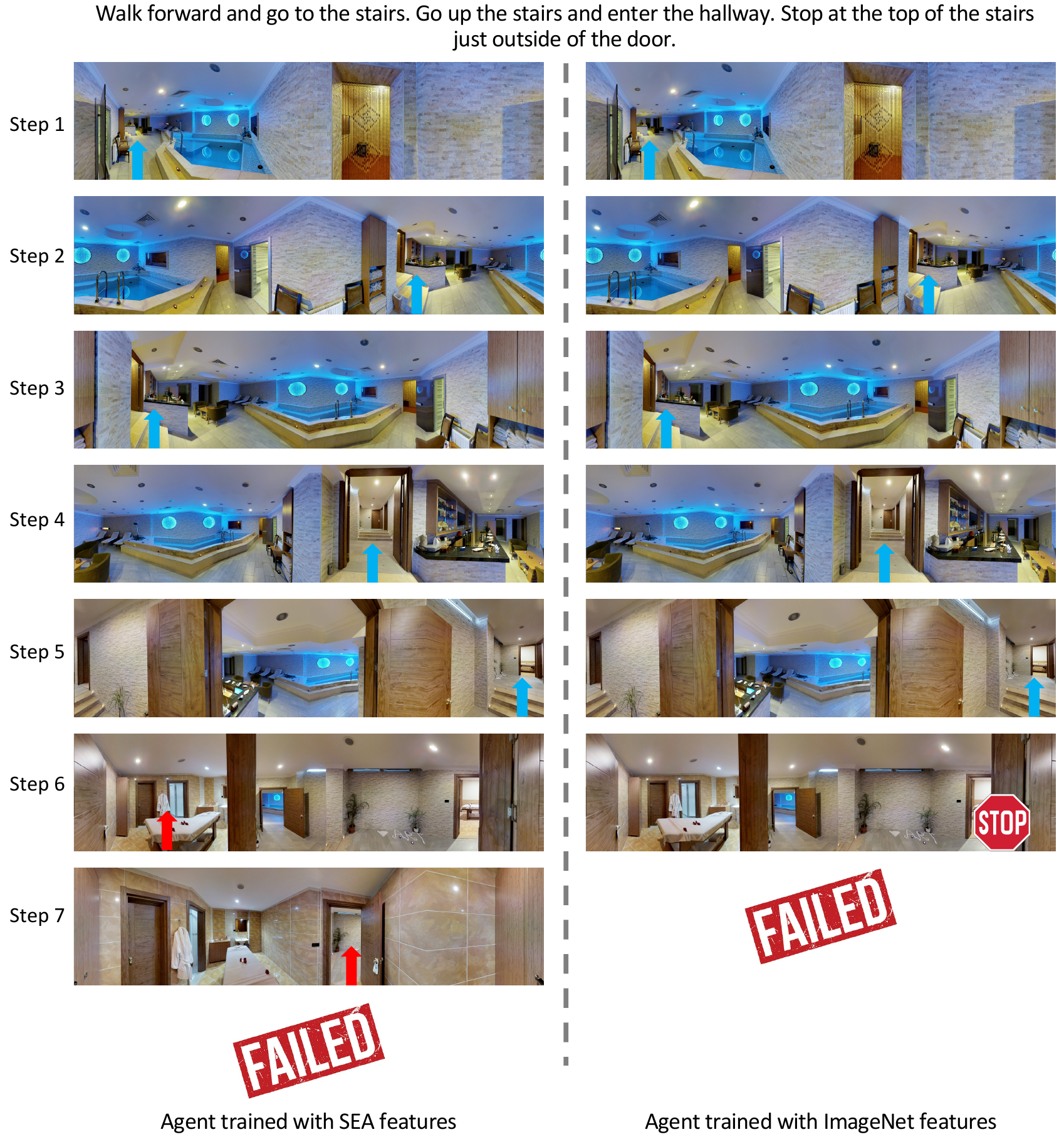}
\caption{
Step-by-step navigation results where both the agents trained with SEA pre-trained features and ImageNet pre-trained features fail.
Both agents fail at the last step of ``\textit{``stop at the top of the stairs just outside of the door''}'' as there are multiple doors at the top of the stairs.
}
\label{supp:fig:sea0im0}
\end{figure*}

We show step-by-step navigation results, where (1) the agent trained with the SEA pre-trained features succeeds while the agent trained with the ImageNet pre-trained features fails in Figure~\ref{supp:fig:sea1im0:object}, and (2) both the agents trained with the SEA pre-trained features and ImageNet pre-trained features fail in Fig.~\ref{supp:fig:sea0im0}.

To successfully follow the instruction of ``\textit{Walk to the oven, turn left, walk past the couch, enter the room past the fireplace, wait at the grey couch.}'' as in Figure~\ref{supp:fig:sea1im0:object}, the agent needs to encode the object information of oven, couch, etc, as well as the spatial information of turn left, walk past, etc.
As can be seen from the figure, the agent trained with ImageNet pre-trained features takes correct actions at the initial steps.
However, it fails at the step of ``\textit{enter the room past the fireplace}''.
The reason could be that the agent fails to interpret the spatial and navigation information of ``\textit{enter the room}'' and ``\textit{past the fireplace}''.
On the other hand, the agent trained with SEA pre-trained features successfully identifies the door, enters the room, and stops at the goal location.

In Fig.~\ref{supp:fig:sea0im0}, both the agents trained with the SEA pre-trained features and ImageNet pre-trained features fail to correctly interpret the instruction of \textit{``stop at the top of the stairs just outside of the door''}.
As can be seen from the panorama at step 6, there are several doors at the top of the stairs.
For the agent trained with SEA pre-trained features on the left, we did not show the complete action sequence since it is trapped in a cycle, where the agent randomly goes toward one of the doors at the top of the stairs.
For the agent trained with ImageNet pre-trained features, it stops at one of the wrong doors at the top of the stairs.

\clearpage

\clearpage

\clearpage
{\small
\bibliographystyle{ieee_fullname}
\bibliography{section/egbib}
}

\end{document}